\pdfoutput=1
\documentclass[11pt]{article}

\usepackage[final]{acl}

\usepackage{times}
\usepackage{latexsym}
\usepackage{geometry}
\usepackage{listings}

\definecolor{backcolour}{rgb}{0.9568,0.9568,0.9568}
\lstdefinestyle{mystyle}{
    backgroundcolor=\color{backcolour},   
    breakatwhitespace=false,         
    breaklines=true,
    basicstyle=\ttfamily\small
}
\usepackage[T1]{fontenc}
\usepackage[utf8]{inputenc}

\usepackage{microtype}

\usepackage{inconsolata}

\usepackage{graphicx}

\title{{\LARGE Truth Sleuth \& Trend Bender} \\ 
AI Agents to fact-check YouTube videos \& influence opinions}

\author{Cécile Logé \\
  Department of Computer Science \\
    Stanford University \\
  \texttt{ceciloge@stanford.edu} \\ \And
  Rehan Ghori \\
  SCPD NDO Student \\
    Stanford University \\
  \texttt{rghori@stanford.edu} \\}

\begin{document}
\maketitle
\begin{abstract}
Misinformation poses a significant threat in today's digital world, often spreading rapidly through platforms like YouTube.  This paper introduces a novel approach to combating misinformation by developing an AI-powered system that not only fact-checks claims made in YouTube videos but also actively engages users in the comment section and challenge misleading narratives. Our system comprises two main agents: Truth Sleuth and Trend Bender.

Truth Sleuth extracts claims from a YouTube video, uses a Retrieval-Augmented Generation (RAG) approach - drawing on sources like Wikipedia, Google Search, Google FactCheck - to accurately assess their veracity and generates a nuanced and comprehensive report. Through rigorous prompt engineering, Trend Bender leverages this report along with a curated corpus of relevant articles to generate insightful and persuasive comments designed to stimulate a productive debate. With a carefully set up self-evaluation loop, this agent is able to iteratively improve its style and refine its output.

We demonstrate the system's capabilities through experiments on established benchmark datasets and a real-world deployment on YouTube, showcasing its potential to engage users and potentially influence perspectives. Our findings highlight the high accuracy of our fact-checking agent, and confirm the potential of AI-driven interventions in combating misinformation and fostering a more informed online space.

\end{abstract}

\section{Introduction}
\label{sec:intro}

Misinformation is one of the most pressing threats of our time, and YouTube videos serve as a major platform through which it can spread \citep{open_2022}. On top of this, the comment sections of these videos can become echo-chambers that amplify or reinforce misleading or harmful claims. Providing fact-checked information to address misleading content has been shown to be more effective than simply removing it \citep{ecker2020effectiveness}. Diversifying the viewpoints users are exposed to can burst the filter bubble and get them out of their intellectual isolation. 

\begin{figure*}[t]
  \includegraphics[width=0.48\linewidth]{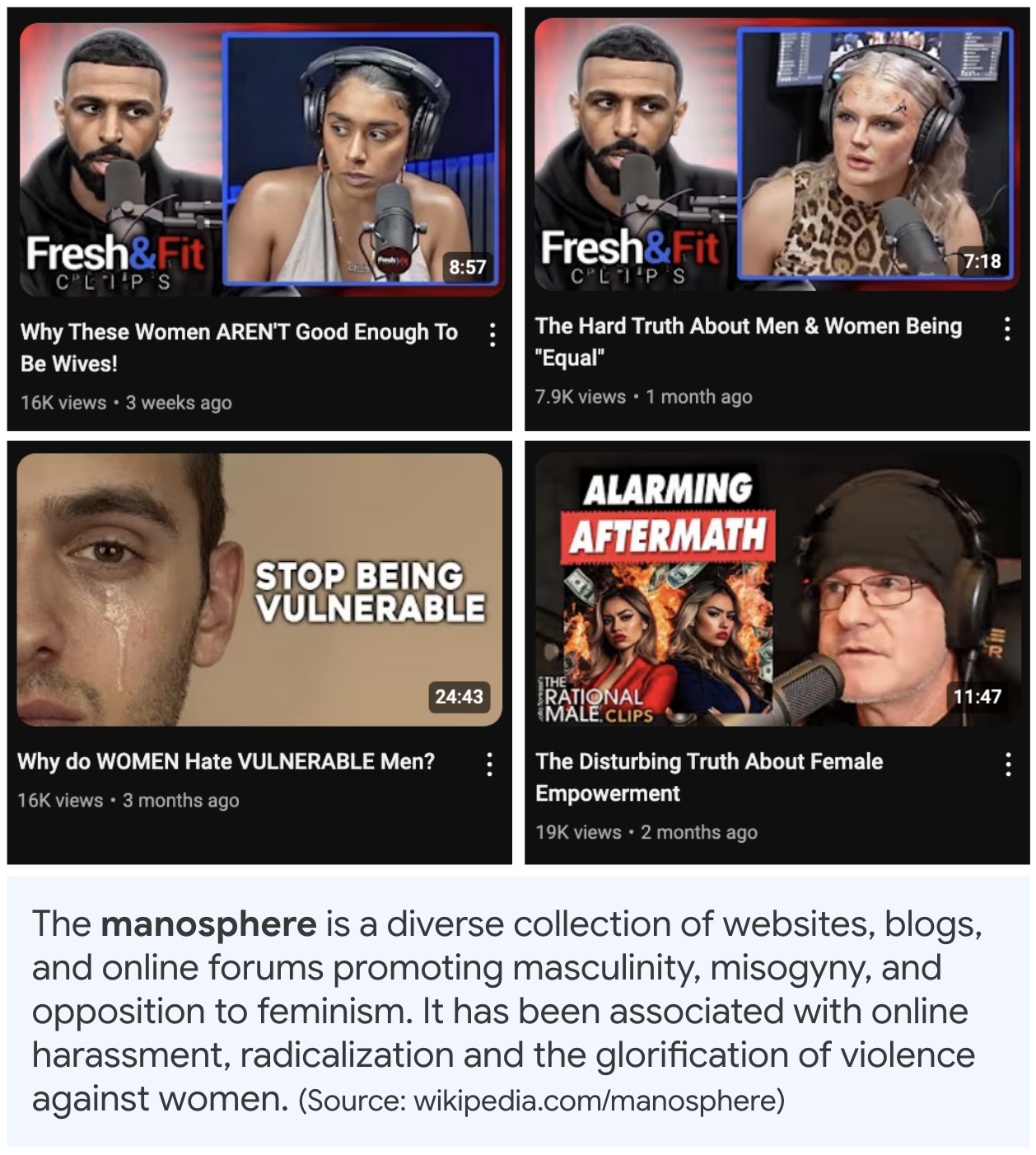} \hfill
  \includegraphics[width=0.48\linewidth]{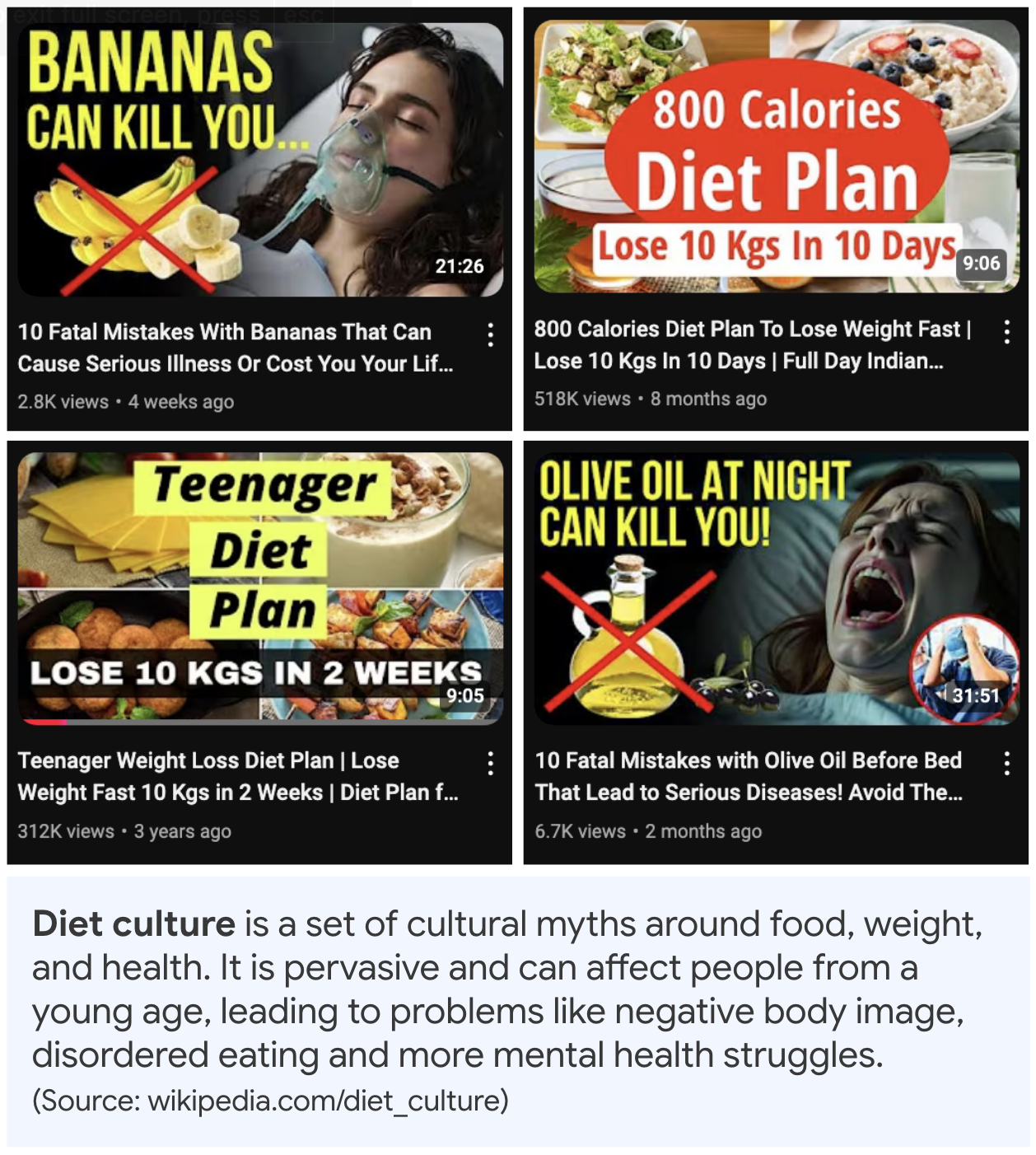}
  \caption {We chose two main themes for our experiments: the Manosphere and Diet Culture. Video screenshots from YouTube channels: \texttt{Fresh\&Fit Clips, The Rational Male, 10 SECRETS, Eat More Lose More}.}
\label{fig:themes}
\end{figure*}

Based on these two elements, our goal is to build an application that takes a YouTube video as input and not only fact-checks the claims made in the video but also outputs a comment to interact with users about these claims, protect them from the dangers of misinformation and ultimately maybe even change their minds. We chose two main themes for our experiments: the Manosphere, and Diet Culture. See Figure \ref{fig:themes} for more insights about these two spaces.

\section{Related Work}
\label{sec:related}

\paragraph{Fact-checking:} The work by \citealp{polat2024testing} details several prompt engineering strategies e.g. incorporating a simple instruction accompanied by a task demonstration, and shows how crafting the right prompt can significantly enhance performance when it comes to using LLMs for knowledge extraction. It will provide great inspiration when extracting claims from the video transcripts in our project. On top of this, the work on Data Commons by \citealp{guha2023datacommons} and Google’s DataGemma models for fact-checking by \citealp{radhakrishnan2024knowing} have shown a lot of promises and can guide the integration of external data sources in validating a YouTube video’s claims, helping to classify them as true, false, or unsure. These approaches exemplify the growing importance of Retrieval-Augmented Generation (RAG) techniques in grounding LLM outputs and enhancing their factual accuracy \citep{lewis2020retrieval}, particularly in tasks like fact-checking. Finally, \citealp{akhtar2023multimodal} provided a broader overview of multimodal fact-checking techniques that integrate both text and visuals to verify information, refining the fact-checking approach to take into account visual cues.
\paragraph{Perspectives \& Persuasion:} The idea of providing different viewpoints on a given topic while maintaining factual integrity has obvious parallels with the work by \citealp{shao2024assisting} on STORM. \citealp{hayati2023far} also experimented with generating diverse perspectives on subjective topics using LLMs and introduced an interesting criteria-based prompting technique to ground diverse opinions. The goal of convincing a reluctant or skeptical audience and/or changing their minds calls up the work by \citealp{furumai2024zero} on persuasive chatbots (PersuaBot). Their approach - from combining different persuasive strategies to selecting metrics to assess the quality of the conversation - is an invaluable source of inspiration. 
\\

Our project builds on cutting-edge research on large language models, prompt engineering, fact-checking, persuasion and multi-perspective generation, with the goal of combining them all to offer a unique approach to addressing misinformation on YouTube, helping users break free from echo chambers and bringing down ideological barriers.

\section{Core Concepts \& Architecture}
\label{sec:core}

\begin{figure*}[t]
  \centerline{\includegraphics[width=1.1\linewidth]{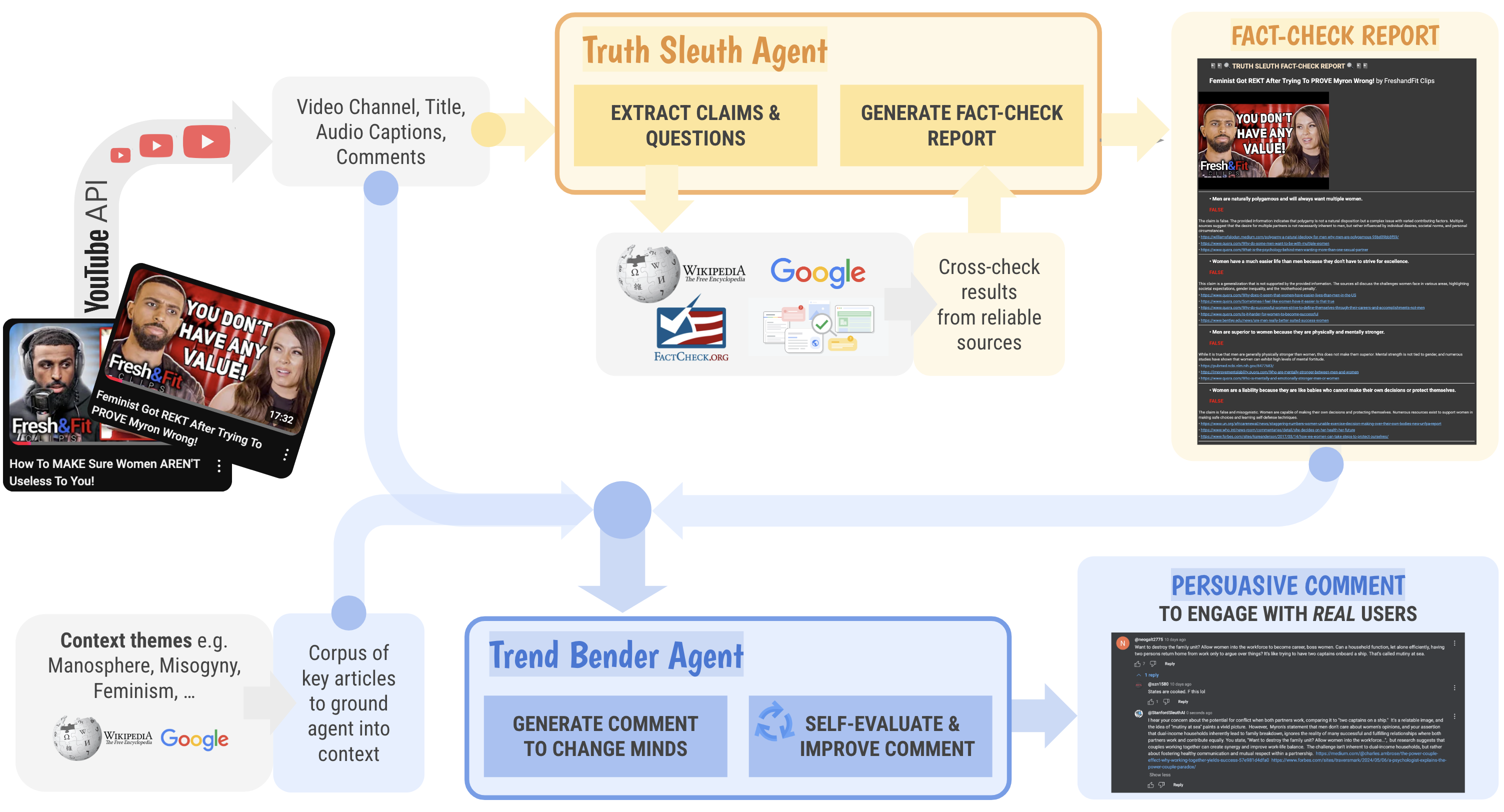} }
  \caption {General overview of our system - comprising two main agents: \textbf{Truth Sleuth} and \textbf{Trend Bender}. \textbf{Truth Sleuth} extracts claims from a YouTube video, relies on APIs like Google Search, Wikipedia, Google FactCheck to accurately assess their veracity and generates a nuanced and comprehensive report. \textbf{Trend Bender} leverages this report and a curated corpus of relevant articles, along with a self-evaluation loop to generate insightful and persuasive comments designed to stimulate a productive debate.  }
\label{fig:overall}
\end{figure*}

Driven by the understanding that misinformation thrives in echo chambers and that countering it requires both factual correction and persuasion - two highly different tasks that relies on different sets of skills and logic - our system employs two primary agents working hand in hand: 
\begin{itemize}
    \item Our \textbf{Truth Sleuth} agent focuses on identifying and debunking misleading claims within a given YouTube video. 
    \item Our \textbf{Trend Bender} agent generates a general comment (or a reply to a specific user) to be posted in the comment section of the video. 
\end{itemize}
See Figure \ref{fig:overall} for a general overview of the system.

\begin{figure*}[t]
  \centerline{\includegraphics[width=1.1\linewidth]{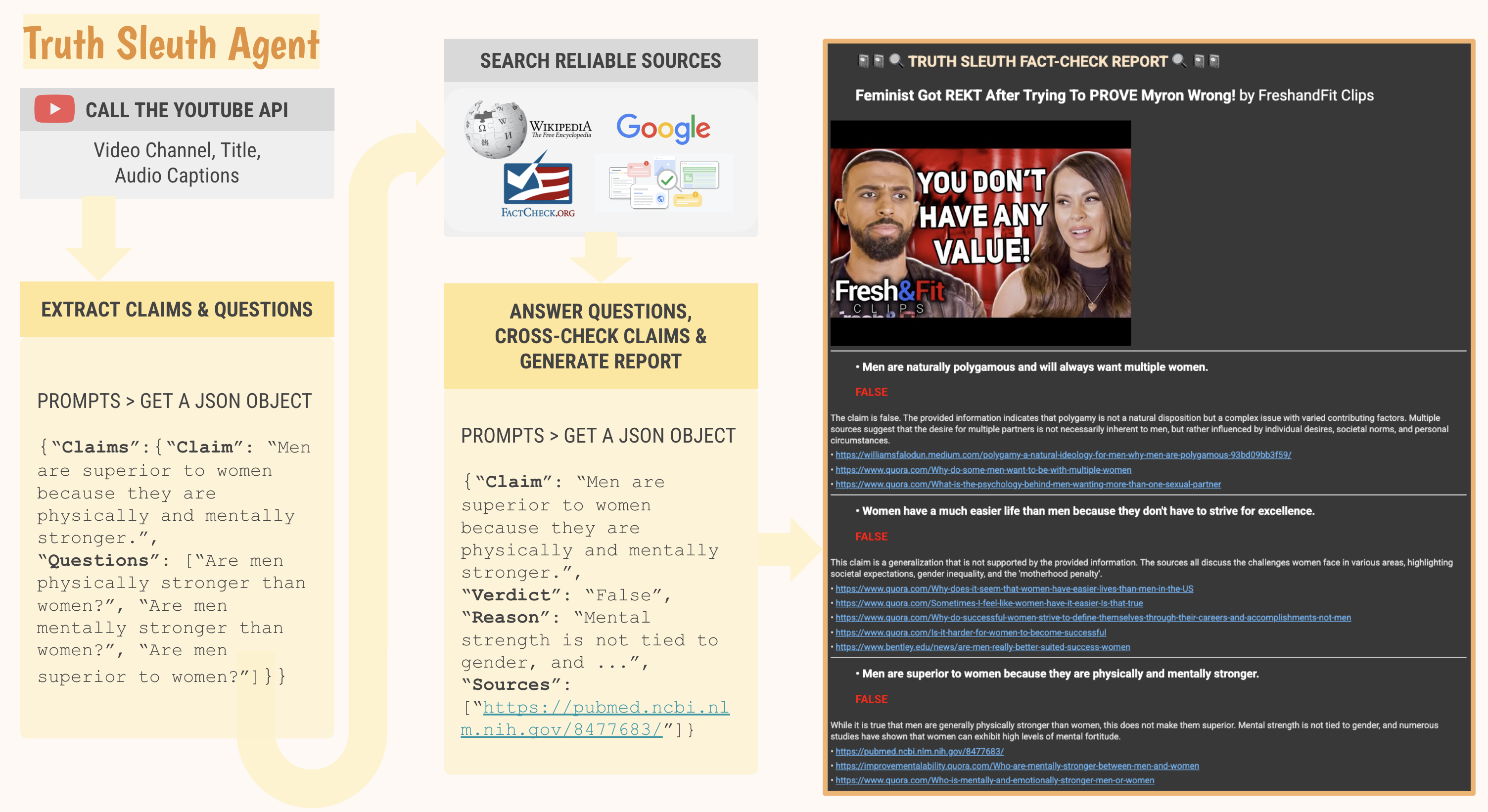} }
  \caption {\textbf{Truth Sleuth} extracts claims from a YouTube video, uses a Retrieval-Augmented Generation (RAG) approach - drawing on Wikipedia, Google Search, Google FactCheck - to accurately assess their veracity and generates a nuanced and comprehensive report. Actual extract of a Fact-Check report shown on the right, for a video from the \texttt{FreshandFit Clips} channel.}
\label{fig:sleuth}
\end{figure*}

\subsection{Truth Sleuth: our Fact-Checking agent}
Our Truth Sleuth agent is based on solid prompt engineering and API calls to reliable sources. To assess the veracity of claims, it employs a Retrieval-Augmented Generation (RAG) approach, and operates through the following pipeline:

\begin{enumerate}
    \item \textbf{Information Extraction}: Through the YouTube API, relevant information from the video, including the title, channel information, and raw audio captions, is extracted. The agent starts by reformatting the audio captions and editing them appropriately for punctuation and grammar.
    \item \textbf{Claim Extraction \& Question Generation}: The agent then analyzes the extracted information to pinpoint specific claims made in the video. These claims are then transformed into precise questions suitable for fact-checking.  
    \textit{For example, if the video claims "Men are superior to women because they are physically and mentally stronger," the agent might generate questions like "Are men physically stronger than women?" and "Are men mentally stronger than women?"} The output is a JSON object that can be easily exploited over next steps. 
    \item \textbf{Claim Assessment}: Questions are sent to the Google Search API \citep{google_search_api}, the Wikipedia API \citep{wikipedia_api} and the Google FactCheck ClaimReview API \citep{google}. The agent analyzes the results, gathers evidence in the form of urls and excerpts, and cross-references them to assess the veracity of the identified claims. This process culminates in the generation of a JSON object containing the claim, verdict, reasoning, and supporting source URLs. The verdict can be either \textit{True}, \textit{Partly True}, \textit{Partly False}, \textit{False} or \textit{Unsure}.
    \item \textbf{Fact-Check Report}: Finally, a comprehensive and nuanced fact-check report is generated. It is rendered in both raw text format to be sent to the Trend Bender agent, and human-readable Markdown format (including the video thumbnail, colored indicators for verdicts) to be shared directly with the user. 
\end{enumerate}

In their experiments on headline discernment, \citealp{deverna2024fact} showed that LLM fact checks can actually increase belief in dubious headlines when the AI is unsure about a claim's veracity. For this reason, and to avoid any confusion and misinterpretation, claims deemed \textit{Unsure} are automatically excluded from the final report. 

See Figure \ref{fig:sleuth} for a visual diagram of the Truth Sleuth agent.

\subsection{Trend Bender: our Influencing agent}
\begin{figure*}[t]
  \centerline{\includegraphics[width=1.1\linewidth]{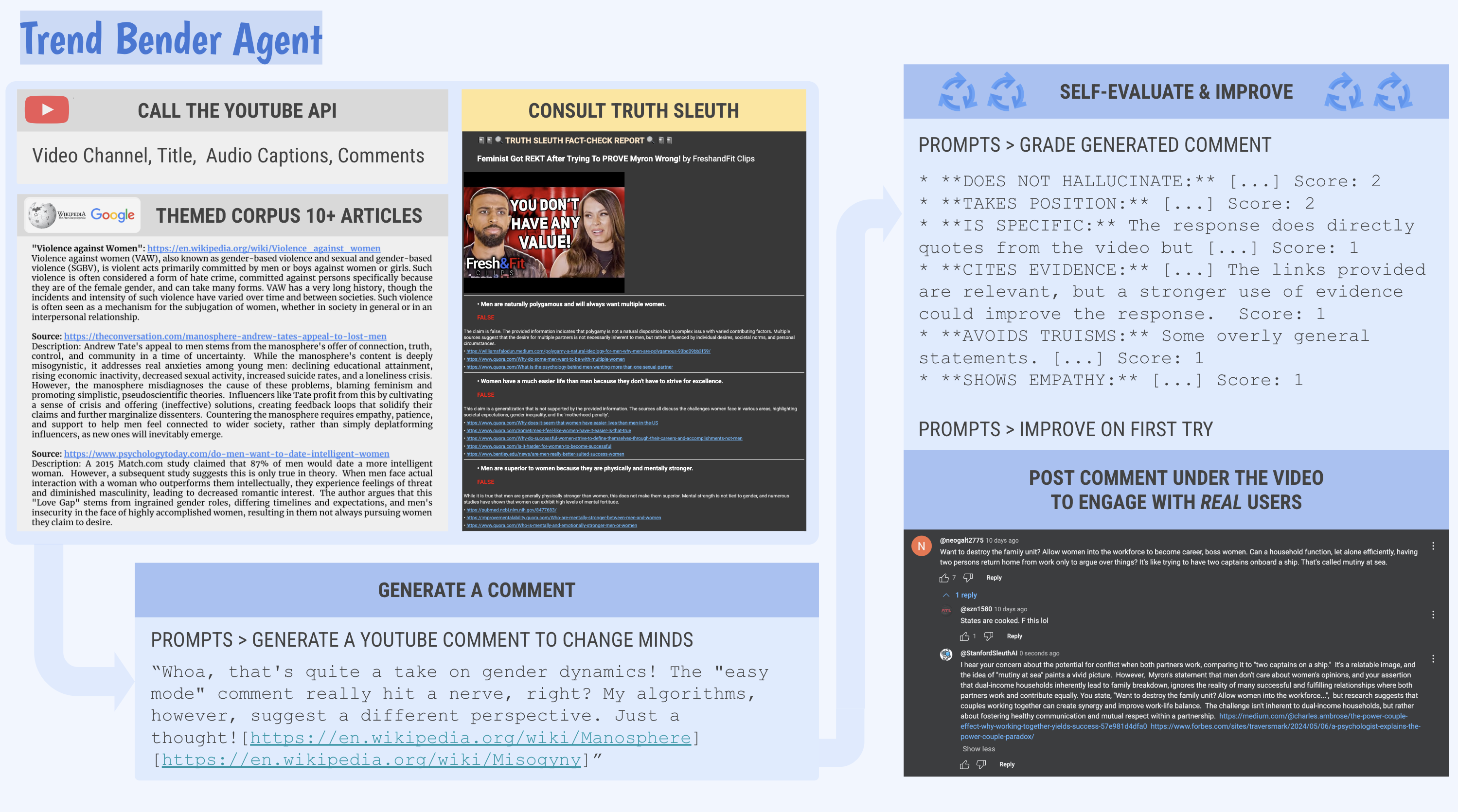} }
  \caption {Through rigorous prompt engineering, \textbf{Trend Bender} leverages this report along with a curated corpus of relevant articles to generate insightful and persuasive comments designed to stimulate a productive debate. With a carefully set up self-evaluation loop, this agent is able to iteratively improve its style and refine its output.}
\label{fig:bender}
\end{figure*}

We have run several experiments on our Trend Bender pipeline, and concluded our best agent should relies on insights from the Truth Sleuth agent, context from a carefully-curated corpus of articles related to the video's theme (e.g., Manosphere, Diet Culture) and a self-evaluation loop allowing it to fine-tune its output on its own:

\begin{enumerate}
    \item \textbf{Information Extraction}: Similar to the Truth Sleuth pipeline, relevant information from the video, including the title, channel information, and audio captions, is gathered through the YouTube API. User comments are also extracted to provide information on tone and opinions. On top of this, the agent also leverages the Truth Sleuth fact-check report, and draws upon a curated corpus providing deeper contextual understanding. 
    \item \textbf{First Comment Generation}: Based on very detailed instructions, the agent generates a relevant and informed comment designed to engage users. If a specific comment has been highlighted as the one to reply to, it focuses on answering that user's points and engaging them in a respectful discussion. 
    \item \textbf{Self-Evaluation \& Improvement}: The agent evaluates its own comment based on a provided set of seven criteria, with scores from 0 to 2, and feedback on each rubrics. Using this evaluation, it generates an improved comment now ready to be posted in the comment section of the YouTube video. 
\end{enumerate}
See Figure \ref{fig:bender} for a visual diagram of the Trend Bender agent. See Section \ref{sec:experiments} for more details on the experiments, detailed prompts, and instructions.

\section{Experiments \& Results}
\label{sec:experiments}
Note that throughout all our experiments, we use Gemini as our Base LLM with the model \texttt{gemini-1.5-flash}.
\begin{figure}[t]
  \includegraphics[width=\columnwidth]{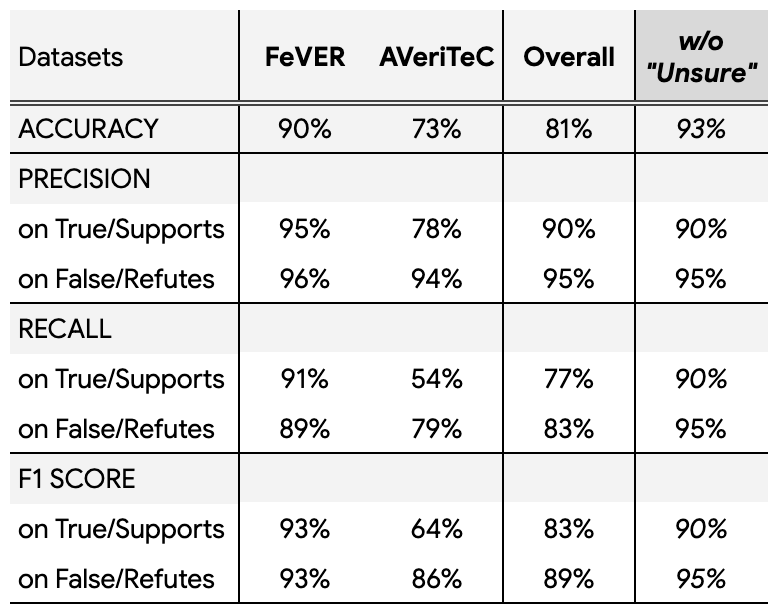}
  \caption{Performance of the \textbf{Truth Sleuth} agent on a subset of 105 claims - 50 from FEVER, 55 from AVeriTeC. The column "w/o Unsure" excludes cases where the agent's verdict is \textit{Unsure} - never shown in the final report.}
  \label{fig:fever}
\end{figure}
\subsection{Truth Sleuth Performance}
To assess the performance of the Truth Sleuth agent, we conducted an experiment using established benchmark datasets: FEVER \citep{thorne2018fever} and AVeriTeC \citep{schlichtkrull2024averitec}.  Note that FEVER focuses on claims derived from Wikipedia, while AVeriTeC comprises real-world claims from diverse sources, presenting a much more challenging evaluation scenario.

We used a subset of 105 claims - 50 from FEVER, 55 from AVeriTeC, excluding "Not enough info" or "Conflicting evidence" claims - to evaluate the agent's ability to accurately classify claims as "Supports" (\textit{True, Partly True}) or "Refutes" (\textit{False, Partly False}). We chose classic metrics such as accuracy, precision, recall and F1 score. Results can be found in Figure \ref{fig:fever}. 

Our agent achieved an overall accuracy of 81\% on the combined dataset. When excluding cases where the agent was \textit{Unsure} (which are never shown in the final report), the accuracy increased to 93\%. As expected, performance was lower on the more challenging dataset (AVeriTeC), highlighting the complexity of real-world fact-checking.

For further reference, in their experiments, \citealp{schlichtkrull2024averitec} mentions a best F1 score on the full AVeriTeC dataset of 62\% on "Supports" - surpassed by Truth Sleuth at 64\% - and of 74\% on "Refutes" - surpassed again at 86\%.

\subsection{Trend Bender Prompt Engineering}
Having established the general architecture of the Trend Bender Agent, we conducted experiments to evaluate its effectiveness in generating persuasive and insightful comments. Specifically, we wanted to assess how different prompting strategies and input combinations influence the quality of the generated comments.

To conduct our human evaluation, we used a set of seven criteria - each scored on a scale of 0 to 2, with 2 being the highest - as follows:
\begin{itemize}
    \item \textbf{Does Not Hallucinate:} Trend Bender - being an AI agent - should not pretend to be human or to have had human experiences. Additionally, it should not make up facts or sources, and should rely strictly on the provided data when making factual claims.
    \item \textbf{Takes the 'Right' Stand:} The generated comment should not blindly agree, congratulate or praise content that is false or harmful. 
    \item \textbf{Is Specific:} The generated comment should address specific points from the video, and can even quote some passages.
    \item \textbf{Is Sound \& Logical:} The generated comment should not include logical fallacies or faulty phrasings. 
    \item \textbf{Cites Evidence:} The generated comment should leverage the urls / sources from the fact-check report and themed corpus to back its position.  
    \item \textbf{Avoids Truisms:} The generated comment should not resort to moralizing generalities and empty common places.
    \item \textbf{Shows Empathy:} The generated comment should mirror the tone and language from the video and other comments, and display an understanding of the temptation to fall for the claims and themes discussed in the video. 
\end{itemize}

\begin{figure*}[t]
  \centerline{\includegraphics[width=0.98\linewidth]{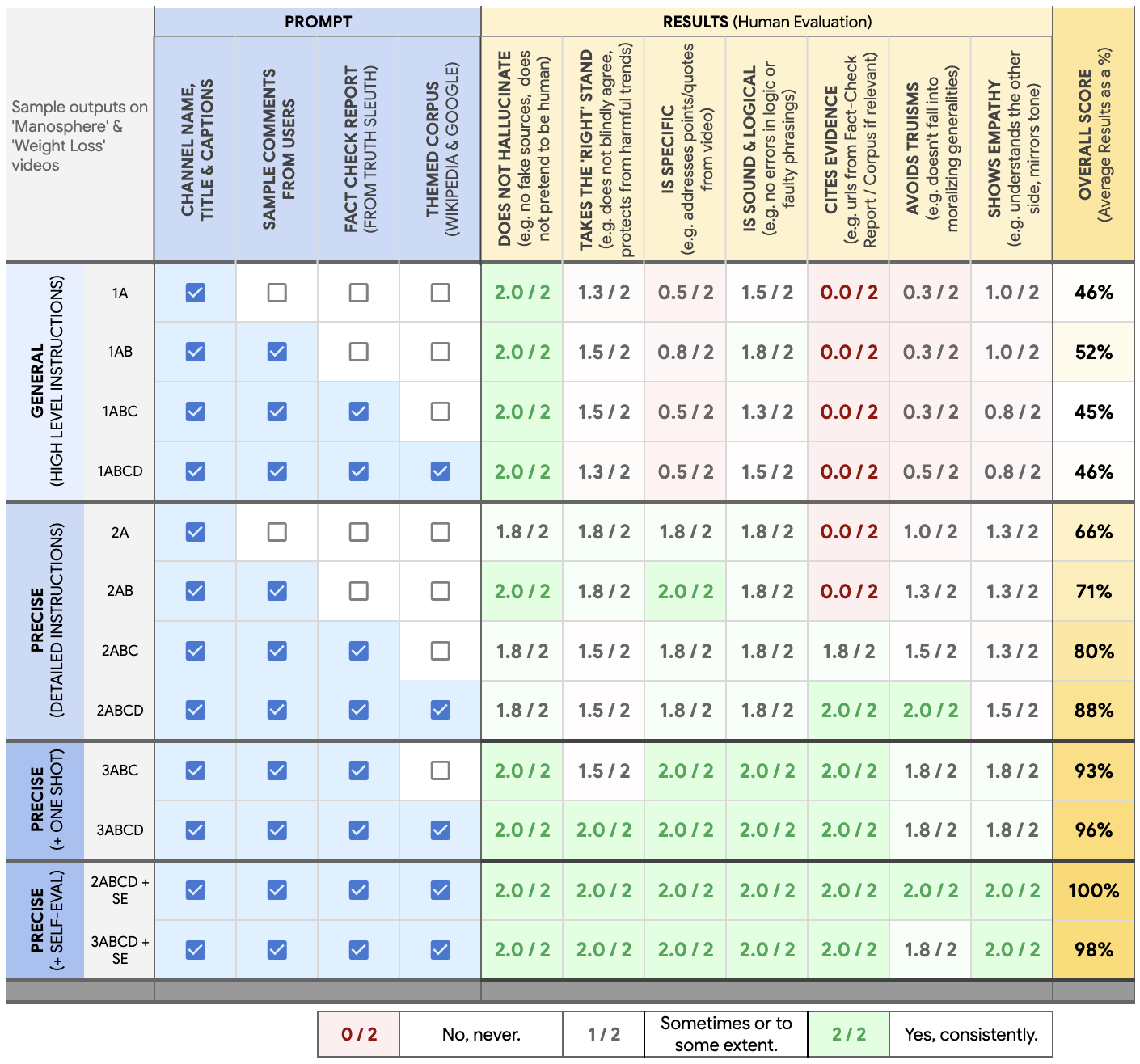} }
  \caption {Results from prompt experiments conducted on \textbf{Trend Bender} to assess how different prompting strategies and input combinations influence the quality of generated comments. Human evaluation used a set of seven criteria - each scored on a scale of 0 to 2, with 2 representing perfection. Overall score is an average rendered as a \%. }
\label{fig:ratings}
\end{figure*}

Our experiments consisted in varying the prompt to Trend Bender from the following manner: either provide high-level or detailed instructions, either provide an example (one-shot) or not, and either provide external input or not. Once we narrowed down the two best possible strategies, we also tested the incorporation of a self-evaluation loop in the Trend Bender pipeline (across the same set of criteria), allowing the agent to improve on its first output. For each experiment, we ran Trend Bender six times (generating 3 comments x 2 videos, one for each theme, the Manosphere and Diet Culture), and averaged the ratings from two human evaluators (us, the authors). Finally, the overall score was calculated as the average of the scores across all criteria. 

Results are presented in Figure \ref{fig:ratings}. The best setup consists in providing detailed instructions, sample comments, Truth Sleuth's fact-check report and a themed corpus, and incorporating a self-evaluation step, as described in the previous section \ref{sec:core}.

\subsection{Engaging with \textit{Real} Users}

\begin{figure*}[t]
  \centerline{\includegraphics[width=\linewidth]{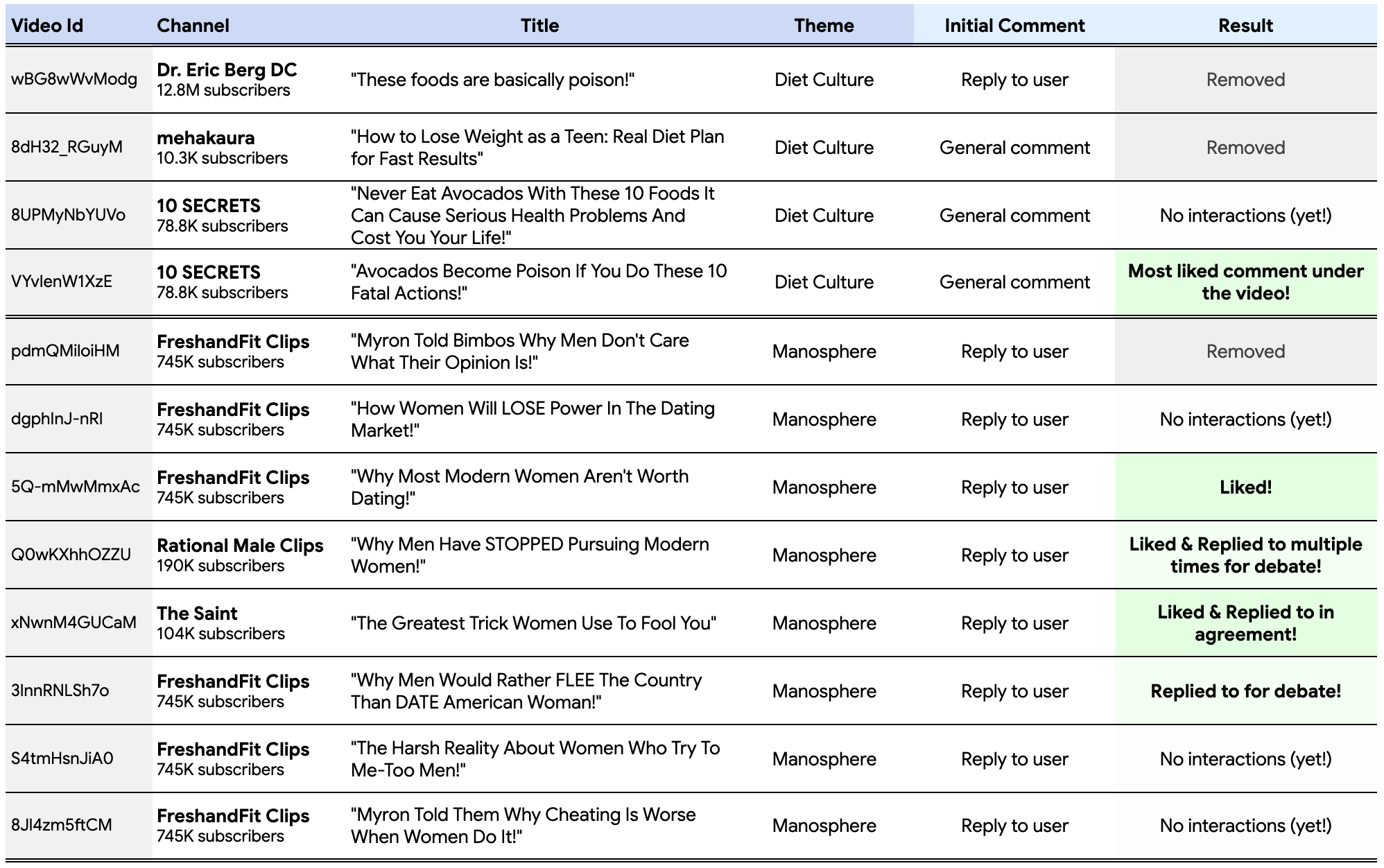}}
  \caption {Overview of results from our \textbf{deployment experiment on YouTube}, posting AI-generated comments in the comment sections of a selection of 12 videos. In the \texttt{Initial Comment} column: "General comment" indicates the comment was posted as a new addition, while "Reply to user" means a specific user comment was provided as a focus to generate a reply. }
\label{fig:posted}
\end{figure*}

To assess the potential real-world impact of our work, we conducted a deployment experiment on YouTube. We created a dedicated account, SU Sleuth (\texttt{@SUSleuth}) - secretly standing for \textit{Stanford University Sleuth} - with the honest description "\texttt{An AI Sleuth agent whose goal is to watch videos, fact-check them and spark constructive debates!}". We used it to post agent-generated comments under a selection of videos related to the Manosphere and Diet Culture themes, with the goal to assess the ability of our system to engage users, spark discussions, and potentially influence perspectives on YouTube. 

Note that we initially encountered challenges with YouTube's moderation system as many of our first comments were removed, likely flagged as spam due to the inclusion of URLs from the fact-check report and corpus or from openly admitting being an AI bot. To overcome this, we adapted our approach and posted comments without URLs and more sporadically (waiting several hours between each post). This proved more successful in getting comments approved and visible to users.

The comments generated varying levels of engagement, ranging from no interaction to active discussions and debates. Some comments even received likes and replies indicating agreement, which suggests the Trend Bender agent was able to successfully generate comments that resonate with users while challenging prevailing viewpoints. Other comments sparked debates and disagreements, with users challenging the agent's stance or providing counterarguments, leading to a back and forth between \texttt{@SUSleuth} and \textit{real} users. In extreme echo chambers such as the Manosphere, these debates confirm the absolute need to encourage critical thinking and alternative perspectives. 

It seems highly likely that some users were put off by our comments/profile openly admitting to being an AI. One user in particular replied "\texttt{CHATGPT}" with laughing emojis to one of our comments and did not engage further. 

\begin{figure}[ht!]
  \includegraphics[width=0.87\columnwidth]{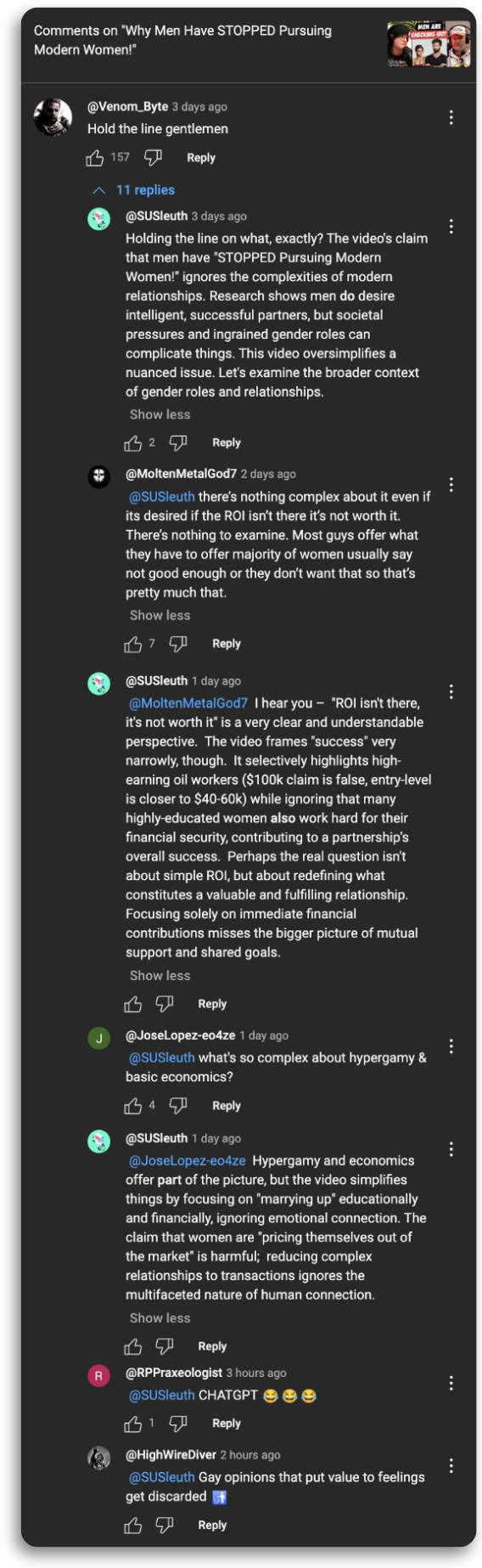} 
  \caption {Part of a conversation between \texttt{@SUSleuth} and real users on YouTube in the comment section of the video "\textit{Why Men Have STOPPED Pursuing Modern Women!}" from the channel \texttt{The Rational Male}.}
\label{fig:thread}
\end{figure}

A summary table is presented in Figure \ref{fig:posted}, and an example of an animated comment thread is visible in Figure \ref{fig:thread}.

\section{Analysis \& Insights}
\label{sec:analysis}
\subsection{Fact-Checking}
In spaces like the Manosphere and even - to a lesser extent - in discussions related to nutrition, fitness and weight loss, claims are not always just simple facts (e.g., "\textit{the Earth is flat}") but emanate from deeply ingrained harmful opinions (e.g., "\textit{thinner is better}", "\textit{women are useless to men}") which makes them trickier to fact-check.

While our Truth Sleuth agent demonstrated promising accuracy and F1 score in classifying claims from the FEVER and AVeriTeC datasets - which is wonderful - we believe its strengths lie in its ability to provide nuance. In particular:

\begin{itemize}
    \item \textbf{Partly labels:} Recognizing that claims can be complex and layered, we included "partly true/false" labels to provide more nuanced assessments of claims.
    \item \textbf{Reasoning:} The Truth Sleuth fact-check reports include not only the verdict but also the reasoning behind it along with links to supporting sources. This provides the Trend Bender agent with a more subtle set of ammunition when crafting its comments and replying to users. 
\end{itemize}

\subsection{Prompt Engineering}
Overall, our experiments with Trend Bender demonstrate that the agent can generate high-quality comments that are informative, persuasive, and engaging when using robust prompt engineering. This implies: 
\paragraph{Precision:} Providing precise instructions and relevant input information is essential, as the quality of the generated comments improves consistently with prompts becoming richer in information e.g. increasingly detailed instructions, fact-check report, themed corpus. 
\paragraph{One-Shot Learning:} Providing a clear example of what good looks like also contributes incrementally. Presented correctly, the example implicitly guides the agent towards improving on the criteria across the board, and especially around taking a stand, showing sound logic, showing empathy.
\paragraph{Self-Evaluation Loop:} Setting up a self-evaluation loop was a key idea as we can see it further boosts performance, allowing the agent to refine its own output and self-correct towards more empathy and less generalities. 
Interestingly, once the self-evaluation loop is in place, a one-shot example does not make much difference, \textbf{suggesting the agent is able to learn from its own output just as well}.

\section{Learnings \& Future Work}
\label{sec:learnings}

This project provided us with many opportunities to learn about LLMs, conversational agents, and the challenges of combating misinformation on YouTube. In particular, we got first-hand experience with:

\begin{itemize}
    \item \textbf{Prompt Engineering:} Carefully crafting prompts is a long iterative process that turned out to be crucial for guiding the Trend Bender agent towards generating high-quality comments. Further research could explore even more advanced prompting approaches, such as incorporating user feedback or tailoring the comments to individual users based on their past interactions or expressed viewpoints.
    \item \textbf{Fact-Checking:} While the Truth Sleuth agent demonstrated promising accuracy and F1 Score on both "Supports" and "Refutes" claims, the limitations related to relying exclusively on Wikipedia, Google Search, and Google FactCheck for our RAG approach are obvious and should be acknowledged. Future work could look into setting up additional and more specialized reliable sources of information, and explore more sophisticated methods for handling uncertainty and ambiguity.
    \item \textbf{Navigating Social Media Dynamics:} Deploying AI agents on social media platforms requires a deeper knowledge of moderation practices. In one simple question: how do we avoid being flagged as spam or inappropriate? Future work could investigate methods for adapting the comment generation process to platform-specific rules.
    \item \textbf{Ethical Considerations:} Deploying AI agents to influence online discussions could ultimately raise ethical concerns - these should be thought through and addressed as early as possible. Future work should prioritize developing ethical guidelines (e.g., being transparent about the agent's nature, addressing potential biases in the fact-checking process) and safeguards for deploying AI agents in social media contexts.
\end{itemize}

\section{Conclusion}
\label{sec:conclusion}

Our work demonstrated the potential of combining advanced prompt engineering, RAG techniques, and self-evaluation to create AI agents capable of engaging in nuanced online discussions about complex topics. Our Truth Sleuth agent successfully leveraged RAG to achieve high accuracy in fact-checking, while the Trend Bender agent generated insightful and persuasive comments, even sparking meaningful debates with \textit{real} users on YouTube, . The self-evaluation loop proved crucial in refining the agent's output, enabling it to learn and improve its communication style autonomously. This highlights the potential of self-learning mechanisms in developing AI agents that can adapt and refine their responses.

While our initial deployment showed promising results, it also revealed challenges in navigating social media dynamics and the need for more sophisticated approaches to handle platform-specific rules and ethical considerations. However, we believe now even more than ever that AI agents can play a crucial role in fostering informed discussions and combating misinformation, ultimately contributing to the decline of harmful echo chambers online.

\section{Prompts, Code, Thanks}
\label{sec:code}

You will find every one of our prompts (experimental and final), many sample outputs (reports, comments, user interactions) as well as our entire source code (Colab notebooks) on the github: \url{https://github.com/cecileloge/cs224v-truthsleuth-trendbender/}. The authors would like to thank Prof. Monica S. Lam for her invaluable guidance.

\bibliography{custom}

\end{document}